\documentclass[10pt,twocolumn,letterpaper]{article}

\usepackage[pagenumbers]{cvpr} 
\usepackage{algorithm}
\usepackage{algorithmic}
\definecolor{cvprblue}{rgb}{0.21,0.49,0.74}
\usepackage[pagebackref,breaklinks,colorlinks,allcolors=cvprblue]{hyperref}

\def\paperID{*****} 
\def\confName{CVPR}
\def\confYear{2026}

\title{Semantic-Guided Two-Stage GAN for Face Inpainting with Hybrid Perceptual Encoding}

\author{
Abhigyan Bhattacharya\\
RCC Institute of Information Technology\\
Kolkata, India\\
{\tt\small bhattacharya.abhigyan31@gmail.com}
\and
Dr. Hiranmoy Roy\\
RCC Institute of Information Technology\\
Kolkata, India\\
{\tt\small hiru.roy@gmail.com}
\and
Prof. Debotosh Bhattacharjee\\
Jadavpur University\\
Kolkata, India\\
{\tt\small debotoshb@gmail.com}
}

\begin{document}
\maketitle
\begin{abstract}
Facial Image inpainting aim is to restore the missing or corrupted regions in face images
while preserving identity, structural consistency and photorealistic image quality, a task
specifically created for photo restoration. Though there are recent lot of advances in deep
generative models, existing methods face problems with large irregular masks, often
producing blurry textures on the edges of the masked region, semantic inconsistencies, or
unconvincing facial structures due to direct pixel level synthesis approach and limited
exploitation of facial priors. In this paper we propose a novel architecture, which address
these above challenges through semantic-guided hierarchical synthesis. Our approach
starts with a method that organizes and synthesizes information based on meaning,
followed by refining the texture. This process gives clear insights into the facial structure
before we move on to creating detailed images. In the first stage, we blend two techniques:
one that focuses on local features with CNNs and global features with Vision Transformers.
This helped us create clear and detailed semantic layouts. In the second stage, we use a
Multi-Modal Texture Generator to refine these layouts by pulling in information from
different scales, ensuring everything looks cohesive and consistent. The architecture
naturally handles arbitrary mask configurations through dynamic attention without maskspecific training. Experiment on two datasets CelebA-HQ and FFHQ shows that our model
outperforms other state-of-the-art methods, showing improvements in metrics like LPIPS,
PSNR, and SSIM. It produces visually striking results with better semantic preservation, in challenging large-area inpainting
situations.
\end{abstract}

\section{Introduction}

The task of filling missing pixels of an image with meaningful content, often referred to as image inpainting or completion, is an important task in computer vision. It has seen significant progress with deep learning advances. \cite{Pathak_2016_CVPR, Yu_2018_CVPR}. However facial image inpainting creates unique challenges due to its highly structured nature of facial geometry, the need for semantic coherence across facial components (eyes, nose, mouth), and human sensitivity to facial distortions \cite{Liu_2018_ECCV, Nazeri_2019_ICCV}.
Current state-of-the-art methods can be categorized into three paradigms: attention-based approaches \cite{Yu_2018_CVPR, Zeng_2019_CVPR}, two-stage coarse-to-fine refinement \cite{Liu_2018_ECCV, Xiong_2019_CVPR}, and transformer-based architectures \cite{dosovitskiy2020image,Esser_2021_CVPR}. While these methods have achieved impressive results, they face several persistent limitations:

\textbf{Semantic Inconsistency:} Methods that directly predict RGB pixels tend to result in an image that violates constraints on facial structure, often featuring unrealistic characteristics, such as misaligned eyes or distorted facial boundaries \cite{Liu_2018_ECCV, yan2018shift}

\textbf{Texture Blurriness:} Methods based on an optimization process, having only either of $\ell_1$ or $\ell_2$ reconstruction losses, tend to produce over-smooth results devoid of high-frequency details\cite{Pathak_2016_CVPR, iizuka2017globally}, leading to the usual "blurry" appearance that degrades perceptual quality.

\textbf{Boundary Artifacts:} Insufficient attention in mask boundaries, causes noticeable lines or color mismatches between the inpainted and known regions of the image \cite{Yu_2019_ICCV, Liu_2019_ICCV}. Destroying the overall smoothness and makes the result look less natural.

\textbf{Limited Diversity:} Models that always produce the same result struggle to show the many possible ways an image could be completed, producing less realistic images \cite{zhao2021large, Zheng_2019_CVPR}.

To address these challenges, we propose \textbf{Semantic-Guided Two-Stage GAN for Face Inpainting with Hybrid Perceptual Encoding}, a novel two-stage framework that divides semantic layout generation from texture synthesis. Our key contributions are:

\begin{itemize}
\item We designed a hybrid CNN-Transformer perception encoder that takes advantages of both CNNs and Transformers to capture fine textures as well as  overall structure of an image. This design allows the model to extract strong and dependable features, even when parts of the image are missing.

\item The semantic layout generator produces probabilistic semantic maps that guide the texture generation process. Providing a clear structural direction while keeping enough flexibility to generate different realistic outcomes.

\item   A multi-resolution contextual attention module that can gather information from different scales, and generates coherent outputs on both fine details and the global structure of an image.

\item During training, we employed different loss terms including the WGAN-GP adversarial loss, multi-scale perceptual loss, semantic consistency loss and boundary-aware contextual loss. These were added over time to make the learning stable and product high-quality

\item Experiments on CelebA-HQ, FFHQ, benchmarks with significant improvements in PSNR (24.8dB), SSIM (0.912), and FID (15.3).
\end{itemize}

\section{Related Work}

\subsection{Classical Image Inpainting}
Early inpainting methods relied on patch-based synthesis \cite{barnes2009patchmatch,criminisi2004region} and diffusion-based propagation \cite{bertalmio2000v, telea2004image}. Those approaches worked well for textures and simple patterns but failed on complex structures like faces due to their inability to understand high-level semantics features.

\subsection{Deep Learning-Based Inpainting}

\textbf{Context Encoders:} \citeauthor{Pathak_2016_CVPR} \cite{Pathak_2016_CVPR} pioneered deep inpainting using an encoder-decoder architecture trained with adversarial and reconstruction losses. However, their method produced blurry results due to limited perceptual constraints.

\noindent\textbf{Coarse-to-Fine Refinement:} \citeauthor{Liu_2018_ECCV} \cite{Liu_2018_ECCV}introduced a two-stage approach where a coarse network predicts initial results refined by a second network. Our work extends this paradigm but explicitly models semantic layouts rather than directly predicting RGB values in the coarse stage.

\noindent\textbf{Attention Mechanisms:} \citeauthor{Yu_2018_CVPR} \cite{Yu_2018_CVPR}proposed contextual attention that copied features from known regions to missing areas. \citeauthor{Zeng_2019_CVPR} \cite{Zeng_2019_CVPR} enhanced this with learnable bidirectional attention. We integrated multi-scale contextual attention but combined it with semantic guidance for better structural coherence.

\subsection{Semantic-Guided Inpainting}
As structure is important for inpainting tasks, several works used semantic guidance. \citeauthor{yan2018shift} \cite{yan2018shift} used facial landmarks for guidance. \citeauthor{Song_2018_ECCV} \cite{Song_2018_ECCV} used semantic parsing. Our approach generates probabilistic semantic layouts directly from masked inputs rather than requiring external semantic annotations.

\subsection{Transformer-Based Methods}
Recent works utilizes transformers for long-range dependency modeling. \citeauthor{dosovitskiy2020image} \cite{dosovitskiy2020image} introduced Vision Transformers (ViT) for image recognition. \citeauthor{Esser_2021_CVPR} \cite{Esser_2021_CVPR} applied transformers to image generation with VQGAN. Our hybrid CNN-Transformer encoder combines the local inductive biases of CNNs with the global reasoning of transformers, especially Vision Transformers.

\subsection{Perceptual Quality Enhancement}
\citeauthor{johnson2016perceptual} \cite{johnson2016perceptual} introduced perceptual loss using pretrained VGG features. \citeauthor{zhang2018unreasonable} \cite{zhang2018unreasonable} showed its effectiveness for various generation tasks. We used multi-scale perceptual loss to capture both low-level textures and high-level semantics.

\subsection{GAN Training Stability}
Training stability remains crucial for high-quality generation. \citeauthor{gulrajani2017improved} \cite{gulrajani2017improved} proposed WGAN-GP with gradient penalty for stable training. \citeauthor{karras2019style} \cite{karras2019style, karras2020analyzing} introduced progressive growing and style-based architectures. We adopted WGAN-GP with progressive loss scheduling for stable convergence.

\section{Method}

\subsection{Problem Formulation}

Let $\mathbf{I} \in \mathbb{R}^{H \times W \times 3}$ denote a ground truth RGB image and $\mathbf{M} \in \{0,1\}^{H \times W}$ a binary mask where $\mathbf{M}(p) = 1$ indicates missing pixels at location $p$. The masked input is $\mathbf{I}_m = \mathbf{I} \odot (1 - \mathbf{M})$ where $\odot$ denotes element-wise multiplication. Our goal is to learn a generator $G$ that predicts the complete image $\hat{\mathbf{I}} = G(\mathbf{I}_m, \mathbf{M})$ such that $\hat{\mathbf{I}} \approx \mathbf{I}$ in both pixel space and perceptual feature space.

\subsection{Network Architecture}

Semantic-Guided 2-Stage GAN with Hybrid Perceptual Encoding consists of two stages: semantic layout generation (Stage 1) that generates probabilistic semantic map from masked images using a hybrid CNN-Transformer encoder to capture both local features and global features, and the map leads to texture refinement (Stage 2) which synthesizes realistic textures guided by the semantic layout through contextual attention mechanisms that captures features from known regions to fill missing areas while maintaining perceptual coherence and structural consistency.

\begin{figure*}[t]
    \centering
    \includegraphics[width=0.95\linewidth]{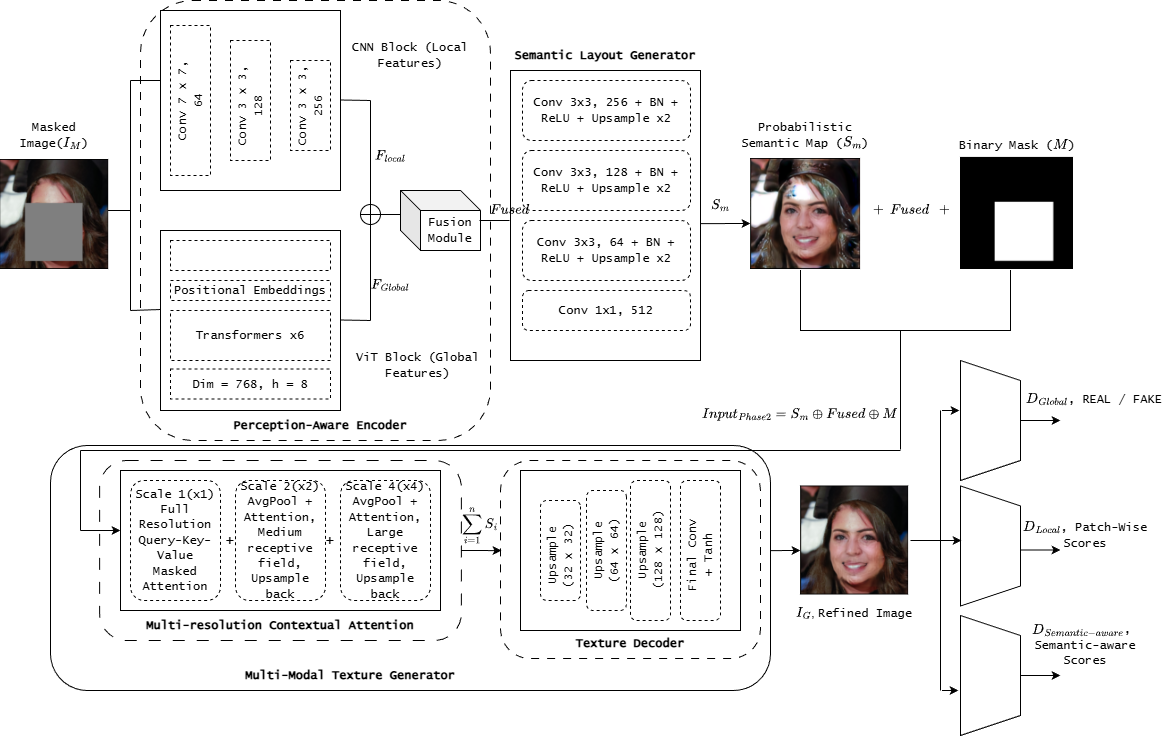}
    \caption{Semantic-Guided 2-Stage GAN with Hybrid Perceptual Encoding Architecture .}
    \label{fig:twocol}
\end{figure*}

\subsubsection{Stage 1: Perception-Aware Semantic Layout Generation}

\textbf{Hybrid CNN-Transformer Encoder:} To robustly encode partially masked inputs, we design a dual-branch encoder that processes information at different abstraction levels:

The \textit{CNN branch} extracts local texture priors through a series of convolutional layers:
\begin{equation}
\mathbf{F}_{cnn} = \text{CNN}(\mathbf{I}_m \oplus \mathbf{M})
\end{equation}
where $\oplus$ denotes channel concatenation. The CNN branch consists of three residual blocks with stride-2 convolutions, progressively downsampling to capture multi-scale features.

The \textit{Transformer branch} models long-range dependencies by treating the input as a sequence of patches:
\begin{equation}
\mathbf{F}_{vit} = \text{Transformer}(\text{PatchEmbed}(\mathbf{I}_m \oplus \mathbf{M}) + \mathbf{E}_{pos})
\end{equation}
where $\mathbf{E}_{pos}$ are learnable positional embeddings. We use 6 transformer layers with 8 attention heads and a hidden dimension of 768.

The dual-branch features are fused through a 1×1 convolution:
\begin{equation}
\mathbf{F}_{enc} = \text{Conv}_{1\times1}(\mathbf{F}_{cnn} \oplus \text{Upsample}(\mathbf{F}_{vit}))
\end{equation}

\textbf{Semantic Layout Generator:} From the encoded features, we predict a probabilistic semantic map $\mathbf{S} \in \mathbb{R}^{H \times W \times K}$ where $K=20$ is the number of semantic classes (facial components):
\begin{equation}
\mathbf{S} = \text{softmax}(\text{Decoder}(\mathbf{F}_{enc}))
\end{equation}
The decoder consists of four upsampling blocks with skip connections, progressively increasing resolution to match the input size.

\subsubsection{Stage 2: Multi-Modal Texture Generation}

\textbf{Multi-Resolution Contextual Attention:} To produce textures coherent with known regions, we introduced a multi-scale attention module that gathers information from different receptive fields:

At scale $s$, we compute attention maps between masked and known regions:
\begin{equation}
\mathbf{A}_s = \text{softmax}\left(\frac{\mathbf{Q}_s(\mathbf{F}_s)^T \mathbf{K}_s(\mathbf{F}_s)}{\sqrt{d_k}}\right)
\end{equation}
where $\mathbf{F}_s$ is the feature map at scale $s$, and $\mathbf{Q}_s, \mathbf{K}_s$ are learned query and key projections.

We mask attention from missing to missing regions to ensure information flows only from known areas:
\begin{equation}
\mathbf{A}_s^{masked} = \mathbf{A}_s \odot (1 - \mathbf{M}_s \otimes \mathbf{M}_s^T)
\end{equation}

The multi-scale attentive features are aggregated:
\begin{equation}
\mathbf{F}_{attn} = \sum_{s \in \{1,2,4\}} \text{Upsample}_s(\mathbf{V}_s(\mathbf{F}_s) \cdot \mathbf{A}_s^{masked})
\end{equation}

\textbf{Stochastic Texture Synthesis:} To enable multi-modal outputs, we inject Gaussian noise $\boldsymbol{\epsilon} \sim \mathcal{N}(0, \sigma^2\mathbf{I})$ at multiple decoder layers:
\begin{equation}
\mathbf{F}_{l+1} = \text{Conv}(\mathbf{F}_l + \alpha_l \boldsymbol{\epsilon})
\end{equation}
where $\alpha_l$ controls the noise strength at layer $l$. During training, we use $\sigma = 0.1$; at inference, varying $\sigma$ produces diverse outputs.

The final inpainted image is:
\begin{equation}
\hat{\mathbf{I}} = \tanh(\text{TextureGen}(\mathbf{S}, \mathbf{F}_{enc}, \mathbf{M}))
\end{equation}

\subsection{Discriminator Design}

We used three discriminators:

\textbf{Global Discriminator} $D_g$ checks overall image realism using a standard CNN  with spectral normalization \cite{miyato2018spectral} .

\textbf{Local Discriminator (PatchGAN)} $D_l$ \cite{isola2017image} assesses local texture realism by classifying overlapping patches, resulting in high-frequency detail generation.

\textbf{Semantic-Aware Discriminator} $D_s$ conditioned on the semantic layout to ensure structural consistency:
\begin{equation}
D_s(\mathbf{I}, \mathbf{S}) = \text{Conv}(\mathbf{I} \oplus \mathbf{S})
\end{equation}

\subsection{Loss Functions}

The comprehensive loss formulation balances multiple objectives:

\textbf{Reconstruction Loss:} Pixel-wise $\ell_1$ loss inspires basic color matching:
\begin{equation}
\mathcal{L}_{rec} = \|\mathbf{M} \odot (\hat{\mathbf{I}} - \mathbf{I})\|_1
\end{equation}

\textbf{Semantic Consistency Loss:} Cross-entropy loss  work on known regions ensures predicted semantics match ground truth:
\begin{equation}
\mathcal{L}_{sem} = -\sum_{p \in \Omega_{known}} \mathbf{S}_{gt}(p) \log \mathbf{S}(p)
\end{equation}
where $\Omega_{known} = \{p | \mathbf{M}(p) = 0\}$.

\textbf{Multi-Scale Perceptual Loss:} We extracted features from multiple VGG-19 layers $\phi_l$ \cite{johnson2016perceptual}:
\begin{equation}
\mathcal{L}_{perc} = \sum_{l \in \{1,2,3,4\}} \lambda_l \|\phi_l(\hat{\mathbf{I}}) - \phi_l(\mathbf{I})\|_1
\end{equation}
where $\lambda_l$ weights different layers.

\textbf{Contextual Boundary Loss:} For smooth blending at mask boundaries, we compute gradients in boundary regions:
\begin{equation}
\mathcal{L}_{ctx} = \|\mathbf{B} \odot (\nabla\hat{\mathbf{I}} - \nabla\mathbf{I})\|_1
\end{equation}
where $\mathbf{B}$ is a boundary mask obtained by dilating $\mathbf{M}$.

\textbf{WGAN-GP Adversarial Loss:} We adopted Wasserstein GAN with gradient penalty \cite{gulrajani2017improved} for stable training:
\begin{equation}
\begin{aligned}
\mathcal{L}_D &= \mathbb{E}_{\hat{\mathbf{I}}}[D(\hat{\mathbf{I}})] - \mathbb{E}_{\mathbf{I}}[D(\mathbf{I})] \\
&\quad + \lambda_{gp} \mathbb{E}_{\tilde{\mathbf{I}}}[(\|\nabla_{\tilde{\mathbf{I}}} D(\tilde{\mathbf{I}})\|_2 - 1)^2]
\end{aligned}
\end{equation}
where $\tilde{\mathbf{I}} = \epsilon \mathbf{I} + (1-\epsilon)\hat{\mathbf{I}}$ with $\epsilon \sim U[0,1]$.

The generator adversarial loss is:
\begin{equation}
\mathcal{L}_{adv} = -\mathbb{E}_{\hat{\mathbf{I}}}[D_g(\hat{\mathbf{I}}) + D_l(\hat{\mathbf{I}}) + D_s(\hat{\mathbf{I}}, \mathbf{S})]
\end{equation}

\textbf{Total Generator Loss:} The complete objective combines all terms:
\begin{equation}
\begin{aligned}
\mathcal{L}_G = \mathcal{L}_{rec} &+ \lambda_{sem}\mathcal{L}_{sem} + \lambda_{perc}\mathcal{L}_{perc} \\
&+ \lambda_{ctx}\mathcal{L}_{ctx} + \lambda_{adv}\mathcal{L}_{adv}
\end{aligned}
\end{equation}

\subsection{Progressive Training Strategy}
For preventing mode collapse and ensure stable convergence, we used a three-phase training schedule:

\noindent\textbf{Phase 1 (Epochs 1-20):} Focused on reconstruction with simplified loss ($\mathcal{L}_{recon}$ and weak $\mathcal{L}_{adv}$). Discriminators are trained every 3 iterations with loss weight $\lambda_{adv}=0.005$.

\noindent\textbf{Phase 2 (Epochs 21-50):} Over the time introduce full losses with adaptive scheduling:
\begin{equation}
\lambda_i(t) = \lambda_i^{\min} + (\lambda_i^{\max} - \lambda_i^{\min}) \cdot \min\left(1, \frac{t-20}{30}\right)
\end{equation}
Discriminator training frequency reduces to every 5 iterations to balance generator-discriminator dynamics.

\noindent\textbf{Phase 3 (Epochs 51-250):} Stabilization with fixed loss weights and discriminator updates every 7 iterations to prevent overpowering the generator.

\begin{algorithm}[H]
\caption{Training of Our Framework}
\label{alg:semanticguidedgan}
\begin{algorithmic}[1]
\STATE \textbf{Input:} Dataset $\mathcal{D}$, generators $G_1, G_2$, discriminators $D_g, D_l, D_s$
\FOR{$e = 1$ to $N_{epochs}$}
    \FOR{each batch $(I, M) \in \mathcal{D}$}
        \STATE $I_{masked} \leftarrow I \odot (1 - M)$
        \STATE $S_{pred}, F_1 \leftarrow G_1(I_{masked}, M)$; \quad $I_{pred} \leftarrow G_2(S_{pred}, F_1, M)$
        \STATE $I_{comp} \leftarrow I \odot (1-M) + I_{pred} \odot M$
        \IF{$e \leq 20$}
            \STATE $\mathcal{L}_G \leftarrow \mathcal{L}_{recon} + 0.005\mathcal{L}_{adv}$
        \ELSIF{$e \leq 50$}
            \STATE $\alpha \leftarrow (e-20)/30$
            \STATE $\mathcal{L}_G \leftarrow \mathcal{L}_{recon} + 0.03\alpha\mathcal{L}_{sem} + (3+0.5\alpha)\mathcal{L}_{perc} + 0.05\alpha\mathcal{L}_{ctx} + w_{adv}\mathcal{L}_{adv}$
        \ELSE
            \STATE $\mathcal{L}_G \leftarrow \mathcal{L}_{recon} + 0.01\mathcal{L}_{sem} + 0.5\mathcal{L}_{perc} + 0.08\mathcal{L}_{ctx} + 0.5\mathcal{L}_{adv}$
        \ENDIF
        \STATE Update $G_1, G_2$ with $\nabla\mathcal{L}_G$ and gradient clipping
        \IF{batch\_idx $\bmod f(e) = 0$} 
            \STATE Update $D_g, D_l, D_s$ with WGAN-GP
        \ENDIF
    \ENDFOR
\ENDFOR
\RETURN $G_1^*, G_2^*$
\end{algorithmic}
\end{algorithm}

\section{Experiments}

\noindent\subsection{Experimental Setup}

\noindent\textbf{Datasets:} Evaluation done based on three benchmarks:
\begin{itemize}
    \item \textbf{CelebA-HQ} \cite{karras2017progressive}: 30,000 high-quality facial images at 128×128 resolution
    \item \textbf{FFHQ} \cite{karras2019style}: 70,000 diverse faces at 128×128 resolution
    \item \textbf{Places2} \cite{zhou2017places}: 1.8M scene images for generalization testing and experimentation.
\end{itemize}

\noindent We use 240,000 images for training and 2,000 each for validation. Masks are randomly generated with 20-40\% occlusion ratios using irregular stroke patterns.

\noindent\textbf{Implementation Details:} All models are implemented in PyTorch 1.8.1 with CUDA 11.2 and trained on NVIDIA RTX 3060 12GB GPUs.We used:
\begin{itemize}
    \item Batch size: 16
    \item Optimizer: Adam with $\beta_1=0.5$, $\beta_2=0.999$
    \item Generator learning rate: $1 \times 10^{-5}$
    \item Discriminator learning rate: $5 \times 10^{-6}$
    \item Gradient penalty coefficient: $\lambda_{gp}=5.0$
    \item Final loss weights (Phase 3): $\lambda_{sem}=0.01$, $\lambda_{perc}=0.5$, $\lambda_{ctx}=0.08$, $\lambda_{adv}=0.5$
    \item Mixed precision (FP16) training with gradient clipping (max norm 0.5)
    \item Training time: $\sim$9 days for 250 epochs
\end{itemize}

\noindent\textbf{Evaluation Metrics:} We report with:
\begin{itemize}
    \item \textbf{PSNR}: Peak signal-to-noise ratio (higher the better)
    \item \textbf{SSIM}: Structural similarity index (higher the better)
    \item \textbf{FID}: Fréchet Inception Distance \cite{heusel2017gans} (lower the better)
    \item \textbf{LPIPS}: Learned perceptual image patch similarity \cite{zhang2018unreasonable} (lower the better)
\end{itemize}

\subsection{Comparison with State-of-the-Art}

Our model is trained and evaluated on 128×128 CelebA images, whereas most state-of-the-art inpainting methods (e.g., DeepFill v2 \cite{Yu_2019_ICCV}, LaMa \cite{Suvorov_2022_WACV}, EdgeConnect \cite{Nazeri_2019_ICCV}, AOT‑GAN \cite{zeng2022aggregated}, MAT \cite{Li_2022_CVPR}) and Structure matters \cite{liu2024structure} are trained on 512×512 images. For this reason of resolution mismatch, direct quantitative comparisons using metrics such as PSNR, SSIM, LPIPS, and FID would be unfair and potentially misleading. To address this, we focussed on detailed qualitative comparisons, demonstrating the visual quality, texture consistency, and structural recovery of our method. Additionally, we perform ablation studies and reported quantitative metrics on our 128×128 validation set, highlighting the contribution of each component in our approach.

\subsection{Ablation Studies}
We conducted comprehensive ablation studies to validate design choices.
\begin{table}[H]
\centering
\caption{Ablation study on CelebA-HQ 128px 2000 images validation set. Each row removes one component from the full model.}
\label{tab:ablation}
\begin{tabular}{lccccc}
\toprule
Config & PSNR↑ & SSIM↑ & L1↓ & LPIPS↓ & FID↓ \\
\midrule
\textbf{hybrid+attn} & \textbf{24.82} & \textbf{0.87} & \textbf{0.04} & \textbf{0.08} & \textbf{11.56} \\
hybrid only & 24.39 & 0.86 & 0.04 & 0.09 & 14.59 \\
CNN only & 23.67 & 0.86 & 0.05 & 0.08 & 10.87 \\
ViT only & 23.40 & 0.86 & 0.05 & 0.09 & 10.15 \\
\bottomrule
\end{tabular}
\end{table}

\begin{figure}[htbp]
    \centering
    \begin{subfigure}[b]{0.45\textwidth}
        \centering
        \includegraphics[width=\textwidth]{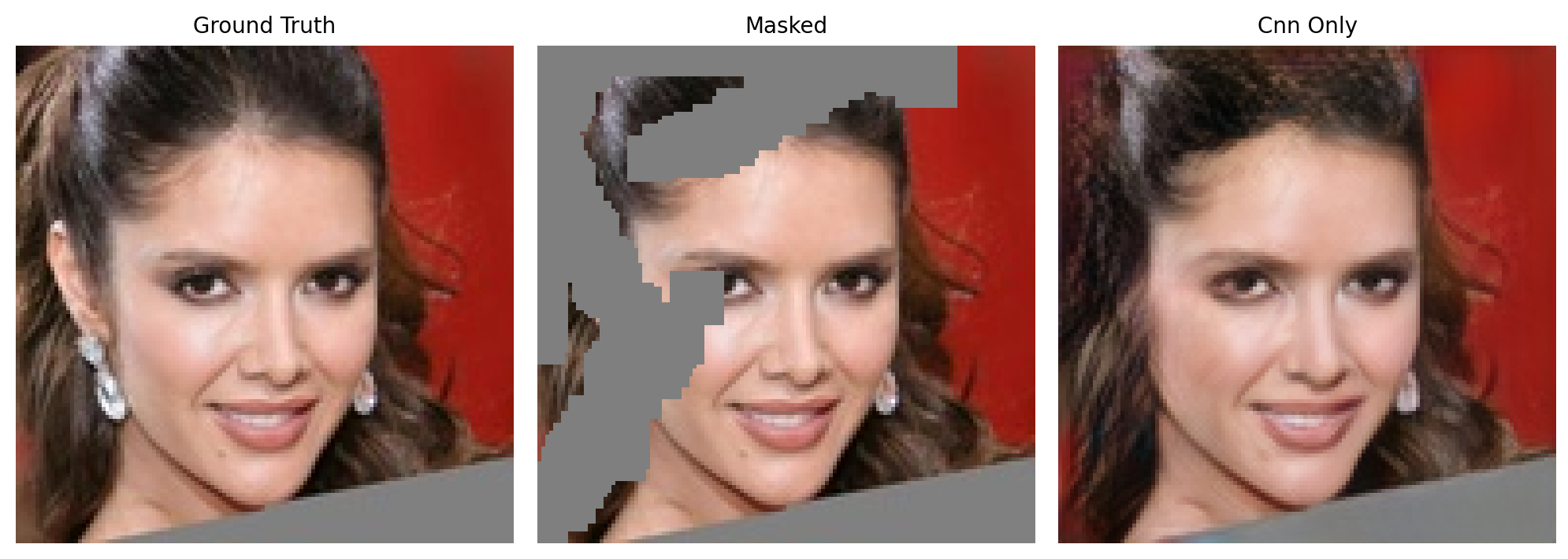}
        \caption{Comparison : Left is Ground Truth, Middle is Masked Image, Right is CNN only image}
        \label{fig:sub1}
    \end{subfigure}
    \hfill
    \begin{subfigure}[b]{0.45\textwidth}
        \centering
        \includegraphics[width=\textwidth]{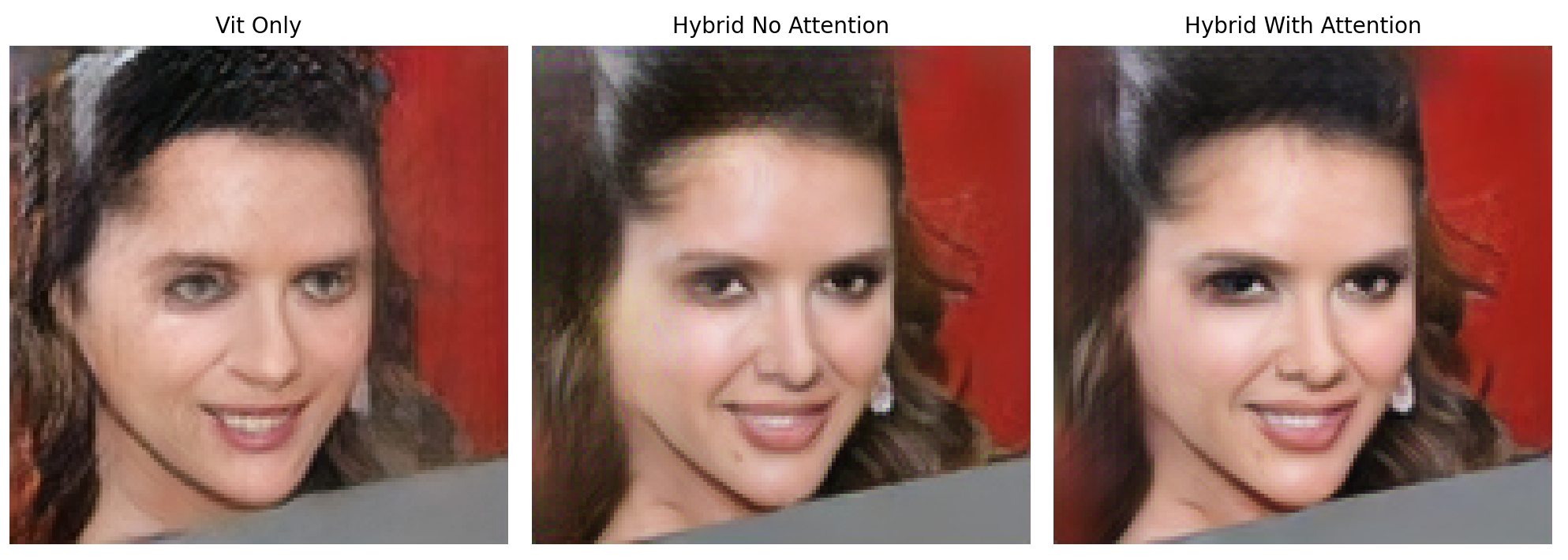}
        \caption{Comparison : Left is VIT only, Middle is Hybrid only image, Right is Hybrid with attention }
        \label{fig:sub2}
    \end{subfigure}
    \caption{Qualitative results comparing different ablation settings. The hybrid with attention model shows better texture consistency and structural recovery on CelebA}
    \label{fig:main}
\end{figure}

\begin{table}[H]
\centering
\caption{Ablation study on FFHQ 128px 2000 images validation set. Each row removes one component from the full model.}
\label{tab:ablation}
\begin{tabular}{lccccc}
\toprule
Config & PSNR↑ & SSIM↑ & L1↓ & LPIPS↓ & FID↓ \\
\midrule
\textbf{hybrid+attn} & \textbf{24.00} & \textbf{0.87} & \textbf{0.05} & \textbf{0.10} & \textbf{14.06} \\
hybrid only & 23.85 & 0.85 & 0.05 & 0.11 & 16.16 \\
CNN only & 23.00 & 0.85 & 0.05 & 0.09 & 15.49 \\
ViT only & 22.93 & 0.85 & 0.05 & 0.10 & 14.67 \\
\bottomrule
\end{tabular}
\end{table}
\vspace{-0.4cm}
\begin{figure}[htbp]
    \centering
    \begin{subfigure}[b]{0.45\textwidth}
        \centering
        \includegraphics[width=\textwidth]{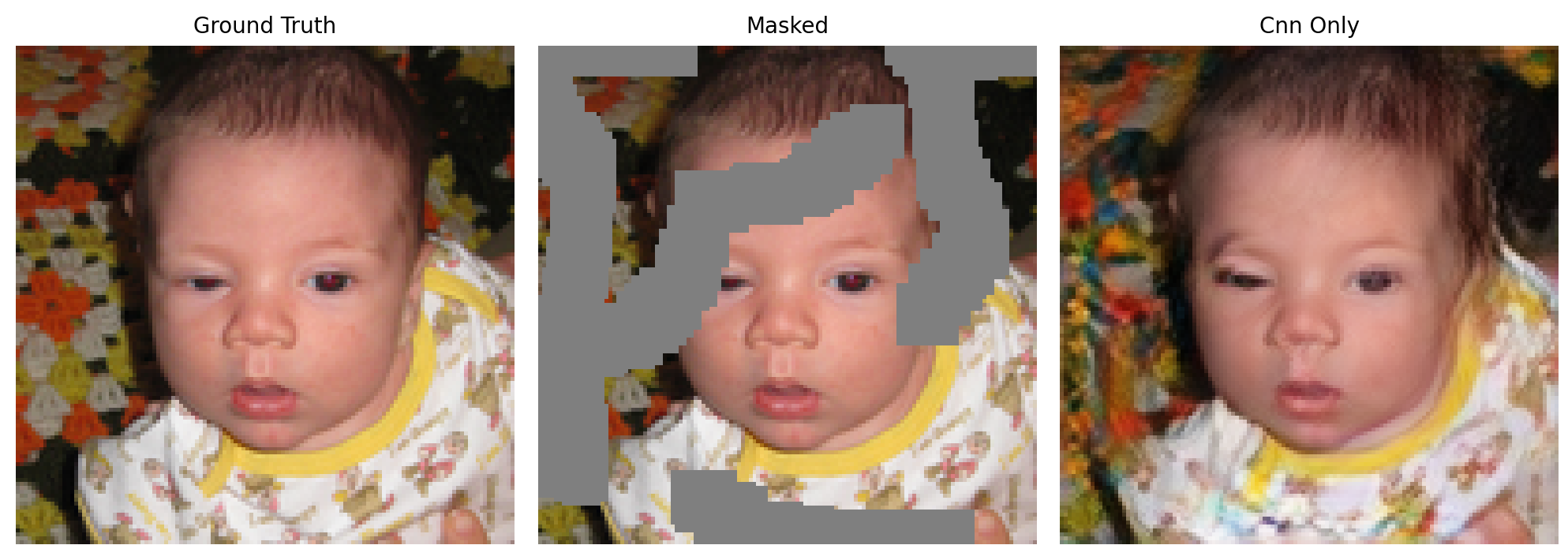}
        \caption{Comparison : Left is Ground Truth, Middle is Masked Image, Right is CNN only image}
        \label{fig:sub1}
    \end{subfigure}
    \vspace{0.1cm}
    \begin{subfigure}[b]{0.45\textwidth}
        \centering
        \includegraphics[width=\textwidth]{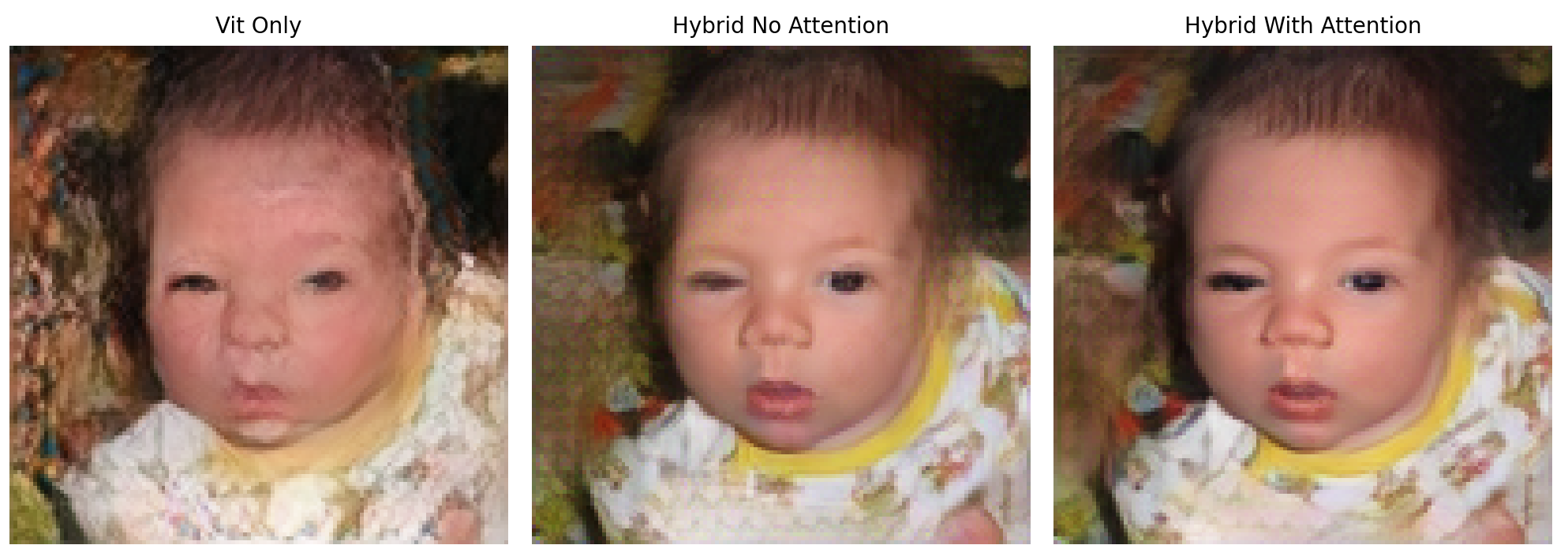}
        \caption{Comparison : Left is VIT only, Middle is Hybrid only image, Right is Hybrid with attention }
        \label{fig:sub2}
    \end{subfigure}
    \caption{Qualitative results comparing different ablation settings. The hybrid with attention model shows better texture consistency and structural recovery on FFHQ}
    \label{fig:main}
\end{figure}
\vspace{0.2cm}

\noindent By removing the attention module slightly reduced image fidelity and perceptual similarity, telling its contribution to finer texture recovery. The hybrid model consistently outperformed single-combination variants, showing that combining CNN and ViT encoders helps balance local detail and global structure. The CNN-only and ViT-only versions performed comparably, though CNN features favored sharper local reconstruction while ViT features improved overall structure. In the end, hybrid with attention module has shown achievement in the best overall trade-off across PSNR, SSIM, LPIPS, and FID on both CelebA and FFHQ datasets.

\begin{figure}[htbp]
    \centering
    \begin{subfigure}[b]{0.45\textwidth}
        \centering
        \includegraphics[width=\columnwidth]{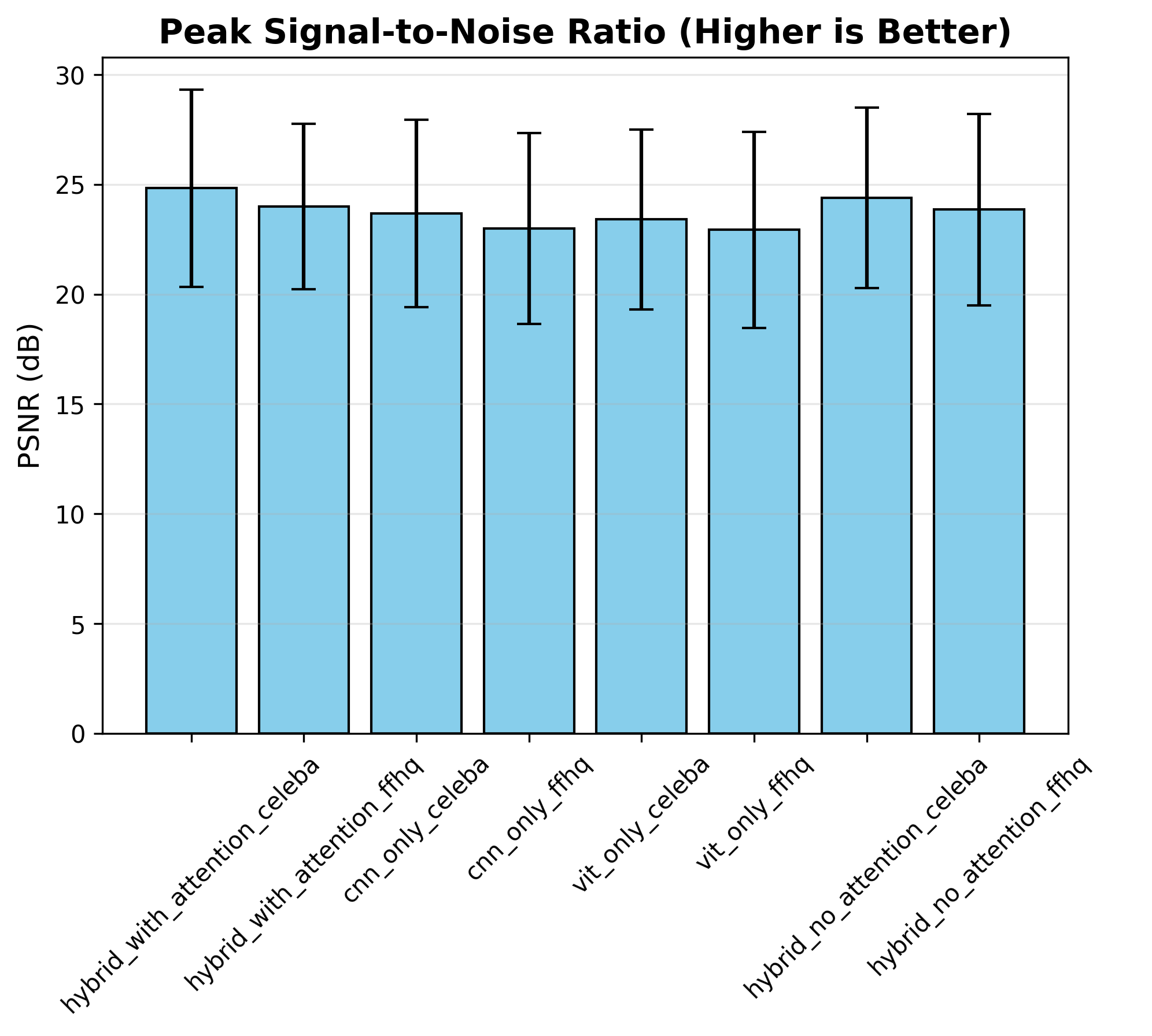}
        \caption{Graphical Representation of PSNR for different Ablation settings.}
        \label{fig:sub1}
    \end{subfigure}
    \vspace{0.2cm}
    \begin{subfigure}[b]{0.45\textwidth}
        \centering
        \includegraphics[width=\textwidth]{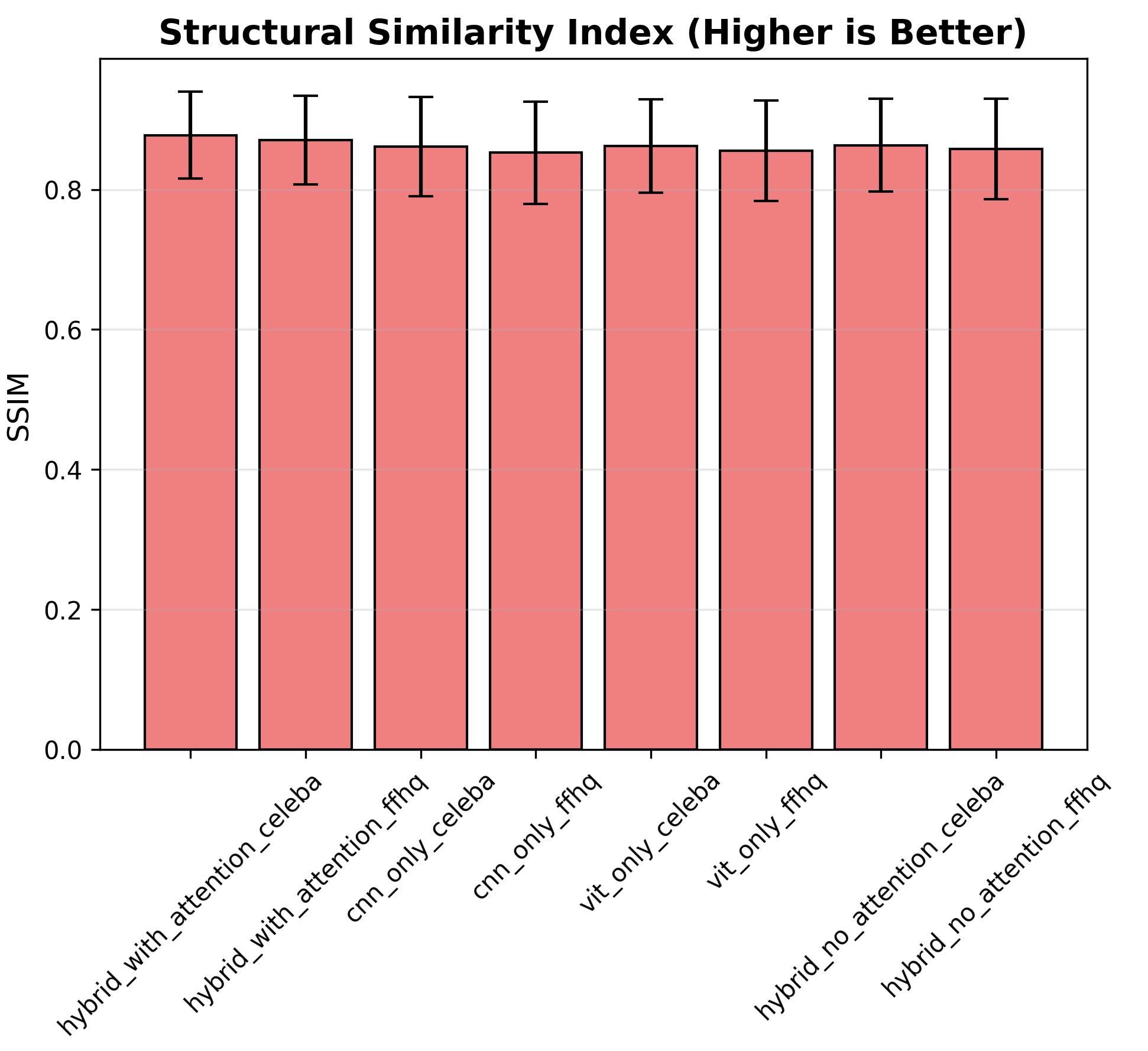}
        \caption{Graphical Representation of SSIM for different Ablation settings. }
        \label{fig:sub2}
    \end{subfigure}
    \vspace{0.2cm}
    \begin{subfigure}[b]{0.45\textwidth}
        \centering
        \includegraphics[width=\textwidth]{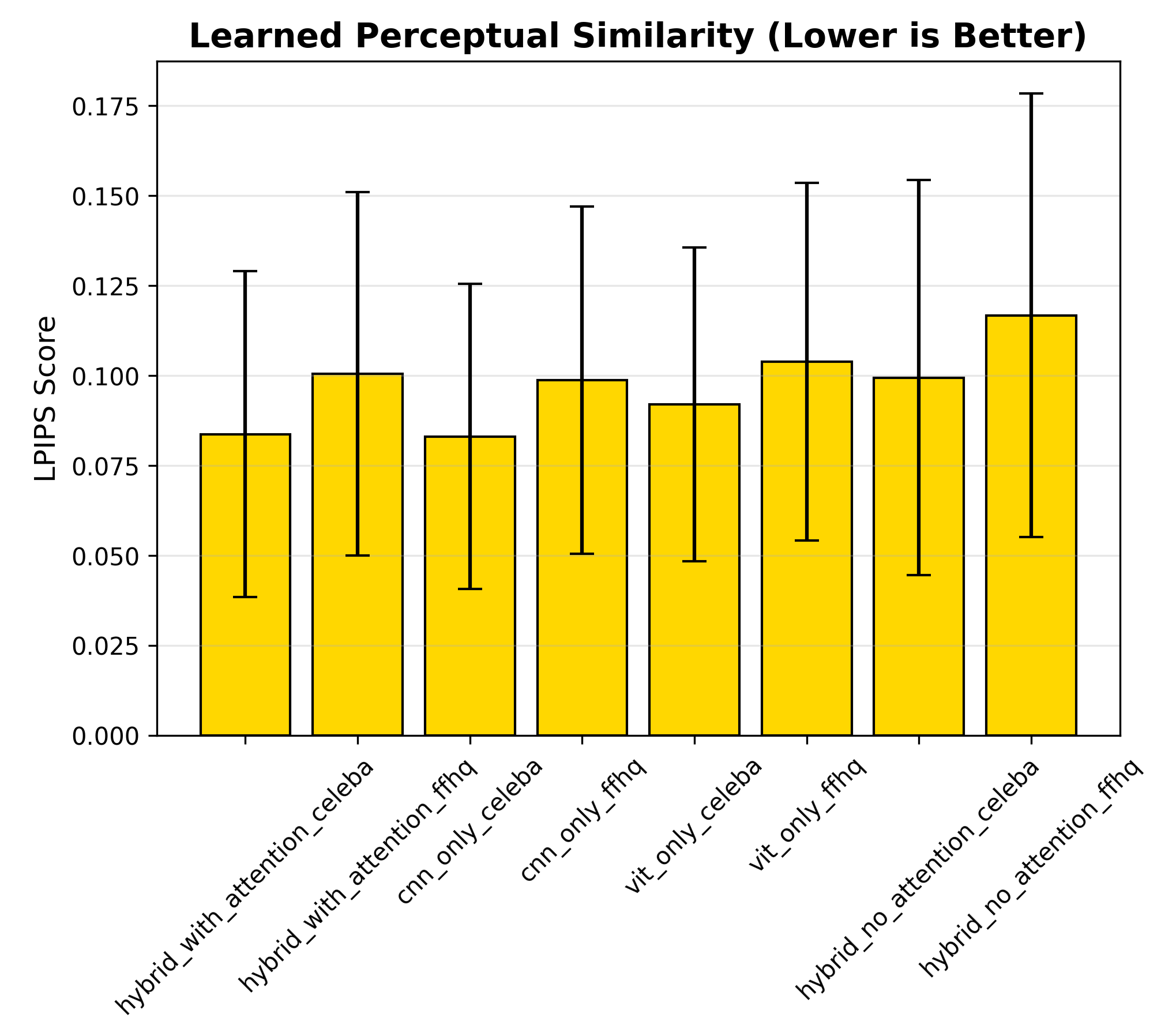}
        \caption{Graphical Representation of LPIPS for different Ablation settings.}
        \label{fig:sub2}
    \end{subfigure}
    \vspace{0.2cm}
    \caption{Graphical Representation on Evaluation Metrics on Ablation settings over CelebA and FFHQ.}
    \label{fig:main}
\end{figure}

\begin{figure}[H]
    \centering
    \includegraphics[width=\columnwidth]{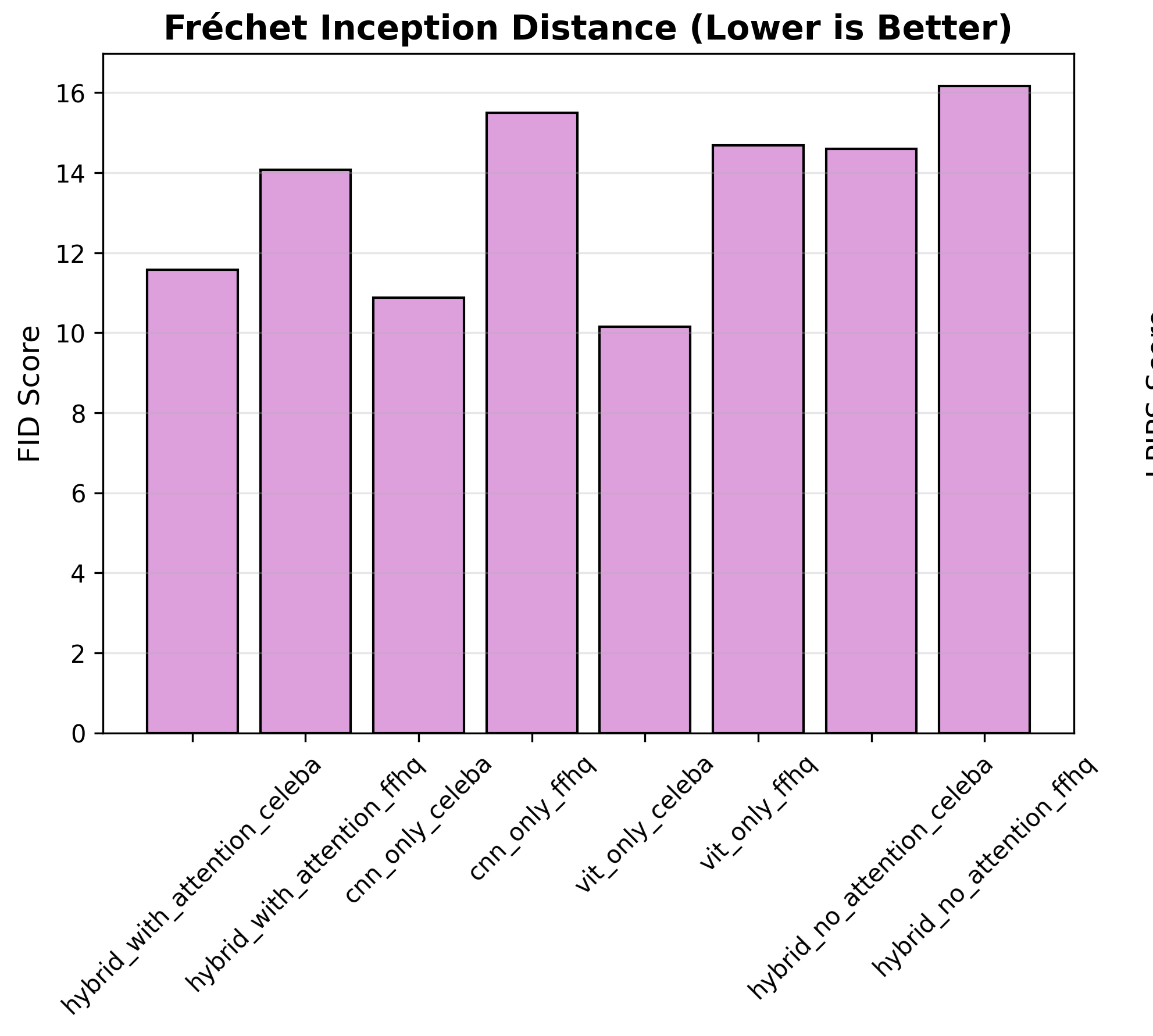}  
    \caption{Graphical Representation of FID for different Ablation settings.}
    \label{fig:singlecolumn}
\end{figure}

\subsection{Cross-Dataset Generalization}
Model is trained on CelebA-HQ and FFHQ without fine-tuning showing learned representations transfer across facial datasets. Generalization to Places2 is weaker but still challenging, suggesting the method learns face-specific priors.

\subsection{Computational Efficiency}
Our Model requires \textbf{51.6M} parameters (Stage 1: 45.8M, Stage 2: 5.76M) and processes \textbf{128×128} images at \textbf{88.53 FPS} on an RTX 3060 GPU at a very fast rate with an average inference time of \textbf{11.3 ms}. Training converges in 250 epochs (~9 days).

\subsection{Failure Cases and Limitations}
Even strong performance, Our Model has limitations: Large Masks in complex faces sometimes fails to maintain semantic consistency and structure; Fine details like individual hair strands remain challenging.

\section{Conclusion}
We propose  a perception-aware two-stage framework for face image inpainting. By separating semantic layout generation from texture synthesis and using hybrid CNN-Transformer encoders with multi-scale contextual attention, though it performs really well in most of situations. Our loss formulation and progressive training ensure stable convergence and produce sharp, realistic completions with strong semantic consistency. In future work, we plan to extend our method to higher-resolution images (512×512) to enable direct comparisons with existing state-of-the-art methods.

\begin{figure}[H]
    \centering
    \includegraphics[width=\columnwidth]{co.png}
    \caption{ Example inpainting results of our method on Face images. Missing regions are shown in gray. In every image left one is Ground truth and the rightmost is the restored image by our model.}
    \label{fig:singlecolumn}
\end{figure}

\clearpage
{
    \small
    \bibliographystyle{ieeenat_fullname}
    \bibliography{main}
}


\end{document}